\documentclass[twocolumn,10pt]{asme2ej}

\usepackage{amsmath}
\usepackage{amsfonts}
\usepackage{amssymb}
\usepackage{bm}
\usepackage{indentfirst}
\usepackage{graphicx}
\usepackage{subfigure}
\usepackage{array}
\usepackage{color}

\DeclareMathAlphabet{\mathsfsl}{OT1}{cmss}{m}{sl}

\newcommand{\PreserveBackslash}[1]{\let\temp=\\#1\let\\=\temp}
\newcolumntype{C}[1]{>{\PreserveBackslash\centering}p{#1}}
\newcolumntype{R}[1]{>{\PreserveBackslash\raggedleft}p{#1}}
\newcolumntype{L}[1]{>{\PreserveBackslash\raggedright}p{#1}}

\definecolor{mygreen}{rgb}{0.1,0.75,0.2}

\numberwithin{equation}{section}

\usepackage{anyfontsize}
\usepackage{enumerate}
\usepackage{esint}
\usepackage{etoolbox}
\usepackage{isomath}
\usepackage{mathrsfs}
\usepackage{mathtools}
\usepackage{multicol}
\usepackage{pgfplots}
\usepackage{scalerel}
\usepackage{stackengine}
\usepackage{tabulary}
\usepackage{tikz}

\makeatletter
\newcommand*\bdot{\mathpalette\bdot@{.65}}
\newcommand*\bdot@[2]{\mathbin{\vcenter{\hbox{\scalebox{#2}{$\m@th#1\bullet$}}}}}
\makeatother

\makeatletter
\newcommand*\bddot{\mathpalette\bddot@{.65}}
\newcommand*\bddot@[2]{\mathbin{\vcenter{\hbox{\scalebox{#2}
    {$\m@th#1\smash{{}_{\bullet}^{\bullet}}$}}}}}
\makeatother

\newcommand{\circled}[2][]{%
  \tikz[baseline=(char.base)]{%
    \node[shape = circle, draw, inner sep = .5pt]
    (char) {\phantom{\ifblank{#1}{#2}{#1}}};%
    \node at (char.center) {\makebox[0pt][c]{#2}};}}
\robustify{\circled}

\makeatletter
\newcommand{\opnorm}{\@ifstar\@opnorms\@opnorm}
\newcommand{\@opnorms}[1]{%
  \left|\mkern-1.5mu\left|\mkern-1.5mu\left|
   #1
  \right|\mkern-1.5mu\right|\mkern-1.5mu\right|
}
\newcommand{\@opnorm}[2][]{%
  \mathopen{#1|\mkern-1.5mu#1|\mkern-1.5mu#1|}
  #2
  \mathclose{#1|\mkern-1.5mu#1|\mkern-1.5mu#1|}
}
\makeatother

\stackMath
\newcommand\reallywidecheck[1]{%
\savestack{\tmpbox}{\stretchto{%
  \scaleto{%
    \scalerel*[\widthof{\ensuremath{#1}}]{\kern-.6pt\bigwedge\kern-.6pt}%
    {\rule[-\textheight/2]{1ex}{\textheight}}
  }{\textheight}%
}{0.5ex}}%
\stackon[1pt]{#1}{\scalebox{-1}{\tmpbox}}%
}

\usepackage{hyperref}   
\hypersetup{
	colorlinks=true,
	linkcolor=blue,
	citecolor=blue,
	urlcolor=blue,
}
\usepackage[sort&compress,square,numbers]{natbib}

\usepackage[normalem]{ulem}
\useunder{\uline}{\ul}{}
\usepackage{graphicx}
\usepackage{amsmath}
\usepackage{lscape}
\usepackage{upgreek}
\usepackage{amssymb}
\usepackage{lineno}
\usepackage{multirow}
\usepackage{pdflscape}
\usepackage{courier}
\usepackage{caption} 
\usepackage{setspace}
\usepackage[utf8]{inputenc}
\usepackage{bm}
\usepackage{mathrsfs}
\usepackage{amsfonts}
\usepackage{color}
\usepackage{float}
\usepackage{import}
\usepackage{url}
\usepackage{multicol}
\usepackage{subfigure}
\newcommand{\real}{\mathbb{R}}
\newcommand{\complex}{\mathbb{C}}

\newcommand{\mcK}{\mathcal{K}}

\newcommand{\mcD}{\mathcal{D}}

\newcommand{\mcG}{\mathcal{G}}

\newcommand{\mcL}{\mathcal{L}}

\usepackage{xcolor}

\def\omg{{\Omega}}

\def \fb{\bm{f}}

\def \ub{\bm{u}}

\def \vb{\bm{v}}
\def \xb{\bm{x}}

\def \hb{\bm{h}}
\def \pb{\bm{p}}

\def \qb{\bm{q}}

\def \cb{\bm{c}}

\newcommand{\vertii}[1]{{\left\vert\left\vert #1
    \right\vert\right\vert}}

\title{A Physics-Guided Neural Operator Learning Approach to Model Biological Tissues from Digital Image Correlation Measurements}

\author{Huaiqian You\\
    {\tensf Graduate Research Assistant}\vspace{10pt}\\
    {\tensfb Quinn Zhang}
    \affiliation{
    Undergraduate Research Assistant \\
	Department of Mathematics \\
	Lehigh University\\
	Bethlehem, PA 18015, USA
    }	
}
\author{Colton J. Ross\\
    {\tensf Graduate Research Assistant}\vspace{10pt}\\
    {\tensfb Chung-Hao Lee}
    \affiliation{
    Assistant Professor \\
	School of Aerospace and Mechanical Engineering \\
	The University of Oklahoma \\
	Norman, OK 73019, USA
    }	
}
\author{Ming-Chen Hsu
    \affiliation{
    Associate Professor \\
	Department of Mechanical Engineering \\
	Iowa State University \\
	Ames, IA 50011, USA
    }	
}
\author{Yue Yu\thanks{Corresponding author.}
    \affiliation{
    Associate Professor \\
	Department of Mathematics \\
	Lehigh University\\
	Bethlehem, PA 18015, USA\\
    Email: yuy214@lehigh.edu
    }	
}

\begin{document}

\maketitle





\begin{abstract}
{\it
We present a data-driven workflow to biological tissue modeling, which aims to predict the displacement field based on digital image correlation (DIC) measurements under unseen loading scenarios, without postulating a specific constitutive model form nor possessing knowledges on the material microstructure. To this end, a material database is constructed from the DIC displacement tracking measurements of multiple biaxial stretching protocols on a porcine tricuspid valve anterior leaflet, with which we build a neural operator learning model. The material response is modeled as a solution operator from the loading to the resultant displacement field, with the material microstructure properties learned implicitly from the data and naturally embedded in the network parameters. Using various combinations of loading protocols, we compare the predictivity of this framework with finite element analysis based on the phenomenological Fung-type model. From in-distribution tests, the predictivity of our approach presents good generalizability to different loading conditions and outperforms the conventional constitutive modeling at approximately one order of magnitude. When tested on out-of-distribution loading ratios, the neural operator learning approach becomes less effective. To improve the generalizability of our framework, we propose a physics-guided neural operator learning model via imposing partial physics knowledge. This method is shown to improve the model's extrapolative performance in the small-deformation regime. Our results demonstrate that with sufficient data coverage and/or guidance from partial physics constraints, the data-driven approach can be a more effective method for modeling biological materials than the traditional constitutive modeling.
}
\end{abstract}

\vspace{0.35 cm}




\section{Introduction}

For many decades, constitutive models based on continuum mechanics have been commonly employed for modeling the mechanical responses of soft biological tissues. In \cite{fung1979pseudoelasticity}, the seminal phenomenological constitutive models were developed, and later employed for the modeling of soft tissues, including the iris \cite{pant2018appropriate}, cardiac heart valves \cite{may1998constitutive,prot2007transversely,sacks2016novel}, arterial vessels \cite{van2011generic}, and the skin \cite{bischoff2000finite}. In the constitutive modeling approaches, a strain energy density function is pre-defined with a specific functional form. Then, the material parameters are calibrated through an inverse method or analytical stress--strain fitting. The descriptive power of these models are often restricted to certain deformation modes/strain ranges, which might lead to limited predictivity and generalizability \cite{lee2014inverse,he2021manifold,lee2017vivo}.

To circumvent such a limitation, data-driven computing has been considered in recent years as an alternative for modeling the mechanical response of biological tissues \cite{pfeiffer2019learning,he2021manifold,he2021deep,tac2021data}. Unlike the traditional material identification techniques in constitutive modeling, data-driven approaches directly integrate material identification with the modeling procedures, and hence do not require a pre-defined constitutive model form. In \cite{pfeiffer2019learning}, a fully convolutional neural network was trained based on synthetic datasets, to estimate a displacement field of material points in the simulated liver organ. In \cite{minano2018wypiwyg}, Mi{\~n}ano et al. constructs the constitutive law for soft tissue damage by solving the system of linear equations consisting of coefficients of shape functions, rather than nonlinear fitting to a pre-defined model. In \cite{he2021manifold}, a local convexity data-driven (LCDD) computational framework was developed that couples manifold learning with nonlinear elasticity, for modeling a representative porcine mitral (heart) valve posterior leaflet's stress--strain data. This framework was further extended to an auto-embedding data-driven approach \cite{he2021deep} to infer the underlying low-dimensional embedding representation of the material database. In \cite{tac2021data}, a neural network was developed to learn the mechanical behavior of porcine and murine skin from biaxial testing data by inferring the relationship between the isochoric strain invariants and the value of strain energy, as well as the strain energy derivatives. Despite these advances, data-driven methods on soft tissue modeling are  mostly focusing on the identification of stress--strain and/or energy--strain relationships for a homogenized material model, and are thus not capable to capture the effects of material spatial heterogeneity. For example, the lack of considering the soft tissue heterogeneity could induce large errors in the predictions of tissue displacements and stresses  \cite{howell2017role}. 

Alternatively, there has been significant progress in the development of deep neural networks (NNs) for heterogeneous material modeling \cite{wang2018multiscale,he2020physics,tartakovsky2020physics,liu2019deep,yang2019derivation,garbrecht2021interpretable,you2022learning,lu2019deeponet,lu2021learning,li2020neural,li2020multipole,li2020fourier}. Among these works, we focus on the neural operator learning approach \cite{you2022learning,lu2019deeponet,lu2021learning,li2020neural,li2020multipole,li2020fourier}, which learns the maps between the inputs of a dynamical system and its state, so that the network serves as a surrogate for a solution operator. Comparing with the classical NNs, the most notable advantage of neural operators is their 
generalizability to different input instances,   
rendering a computing advantage on prediction efficiency -- once the neural operator is trained, solving for a new instance of the input parameter only requires a forward pass of the network. In \cite{yin2022simulating,goswami2022physics,yin2022interfacing}, neural operators have been successfully applied to modeling the unknown physics law of homogeneous materials. In \cite{li2020neural,li2020multipole,li2020fourier,lu2021comprehensive}, neural operators were used as a solution surrogate for the Darcy's flow in a heterogeneous porous medium with a known microstructure field. In our previous work \cite{you2022learning}, an implicit neural operator architecture, namely the implicit Fourier neural operator (IFNO), was proposed to model heterogeneous material responses without using any pre-defined constitutive models or microstructure measurements. In particular, we have investigated the applicability of learning a material model for a latex material directly from digital image correlation (DIC) measurements, and shown that the learned solution operators substantially outperform the conventional constitutive models such as the generalized Mooney--Rivlin model.

To the best of our knowledge, the neural operator learning approaches have not been applied to soft tissue biomechanics. Moreover, the effectiveness of neural operator learning methods in extrapolation to small and large deformation regimes has yet to be systematically examined. To achieve these goals, in this work we propose to advance the current data-driven methods on soft tissue modeling by extending the neural operator learning approach. In particular, we employ the IFNO to learn the material model from DIC measurements on a representative tricuspid valve anterior leaflet (TVAL) specimen from a porcine heart and assess its predictability on unseen and out-of-distribution loading scenarios. To further improve the generalizability of the proposed framework, we also infuse partial physics knowledge via a soft penalty constraint to obtain a novel physics-guided neural operator learning framework. This method is shown to improve the extrapolative performance of our model in the small deformation regime.

\begin{figure*}[!ht]
\centering
\subfigure{\includegraphics[width=1.95\columnwidth]{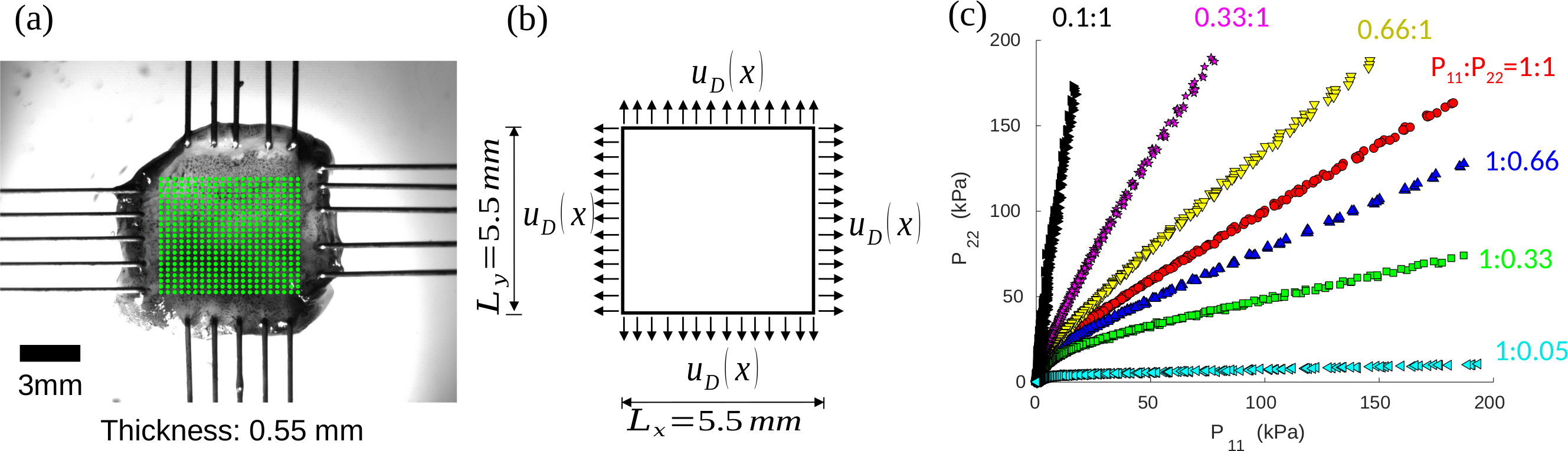}}
\caption{Problem setup for the proposed data-driven computations: (a) an image of the speckle-patterned porcine tricuspid valve anterior leaflet (TVAL) specimen subject to biaxial stretching (the DIC (digital image correlation) tracking grid is shown in green), (b) schematic of a specimen subject to Dirichlet-type boundary conditions, as the corresponding numerical setting of (a), and (c) first Piola-Kirchhoff stresses $P_{11}$ vs. $P_{22}$ of seven biaxial tension and constrained uniaxial tension testing protocols, with further details provided in Table \ref{tab:dic_setting}.}
\label{fig:dicsetup}
\end{figure*}

    \begin{table*}[!ht]
        \centering
        \caption{Protocols of the mechanical testing on a representative TVAL specimen. The resultant displacement fields, based on digital image correlation, were used in the data-driven computations. ($P_{11}$ and $P_{22}$ denote the first Piola-Kirchhoff stresses in the $x$- and $y$-directions, respectively, and $\lambda_1$, $\lambda_2$ are the stretches ratios in these two directions.)}
        {\renewcommand{\arraystretch}{0.95}
        \begin{tabular}{|>{\hspace{-4pt}}c<{\hspace{-4pt}}|>{\hspace{-4pt}}c<{\hspace{-4pt}}|>{\hspace{-4pt}}c<{\hspace{-4pt}}|>{\hspace{-4pt}}c<{\hspace{-4pt}}|>{\hspace{-4pt}}c<{\hspace{-4pt}}|>{\hspace{-4pt}}c<{\hspace{-4pt}}|>{\hspace{-4pt}}c<{\hspace{-4pt}}|}
        \hline
        Set ID& Experiment Protocol   & $\max(\lambda_1)$ & $\max(\lambda_2)$ & $\max(P_{11})$ & $\max(P_{22})$& \# of Samples\\
        \hline
        1& Biaxial Tensions $P_{11}:P_{22}=1:1$ &1.46&1.68&184.1 kPa&165.1 kPa& 3,921 \\
        2& Biaxial Tensions $P_{11}:P_{22}=1:0.66$ &1.48&1.63&187.1 kPa&127.8 kPa& 3,797  \\
        3& Biaxial Tensions $P_{11}:P_{22}=1:0.33$  &1.52&1.52&186.9 kPa&74.1 kPa& 3,539  \\
        4& Biaxial Tensions $P_{11}:P_{22}=0.66:1$ &1.42&1.72&145.9 kPa&188.2 kPa& 4,013  \\
        5& Biaxial Tensions $P_{11}:P_{22}=0.33:1$ &1.32&1.79&77.9 kPa&189.8 kPa& 4,175  \\
        6& Constrained Uniaxial in $x$, $P_{11}:P_{22}=0.05:1$ &1.56&1.0&197.9 kPa&10.6 kPa& 3,539 \\
        7& Constrained Uniaxial in $y$, $P_{11}:P_{22}=1:0.1$ &1.0&1.89&17.2 kPa&176.1 kPa& 3,539  \\
        \hline 
        \end{tabular}}
        \label{tab:dic_setting}
    \end{table*}

The remainder of this paper is organized as follows. In Section \ref{sec:background}, we introduce our data-driven computing paradigm based on the neural operator learning method, which integrates material identification, modeling procedures, and material responses prediction into one unified learning framework. In particular, a stable deep layer architecture, i.e., the IFNO, is introduced in Section \ref{sec:ifno} and incorporated with partial physics knowledge in Section \ref{sec:pi}. In Section \ref{sec:setting}, we introduce our experimental setting on a representative TVAL specimen. Four study scenarios, considering different sets of experimental data for model training and predictions, are used to examine the in-distribution and out-of-distribution predictivity of the proposed IFNO method. The effectiveness of the IFNO approach is also compared with finite element simulation results based on a phenomenological constitutive model. Then, we illustrate the prediction results of the IFNOs and physics-guided IFNOs, and compared their results with the modeling results based on the fitted constitutive model in Section \ref{sec:results}. Finally, we provide a summary of our achieved goals and concluding remarks in Section \ref{sec:conclusion}.

\section{An Integrated Learning Framework}\label{sec:background}

In this section, we first formulate the proposed workflow of data-driven material modeling using the operator learning framework, and then introduce the deep neural operator model -- the implicit Fourier neural operator (IFNO) \cite{you2022learning}. Next, we propose to further infuse partial physics knowledge via a soft penalty constraint to guide the training and prediction of the neural operators.


\subsection{Neural operator learning methods}

The main objective of this work is to model the mechanical response of a representative soft biological tissue directly from DIC-tracked displacement measurements, without any pre-defined constitutive model nor knowledge on the tissue microstructure. As depicted in Figure \ref{fig:dicsetup} and Table \ref{tab:dic_setting}, let us consider a soft biological tissue specimen which is mounted to a biaxial testing system and deforms under external loading. Denoting the region of interests on this specimen as a 2D domain $\Omega$, our aim is to identify the {\it best surrogate operator}, that can accurately predict the displacement field $\ub(\xb)$, $\xb\in\omg$, given new and unseen loading scenarios. In this work, we model the tissue mechanical response as a quasi-static and hyperelastic problem for simplicity, so the resultant displacement field can be fully determined by a displacement-type loading applied on the domain boundary $\partial \Omega$. Thus, given the Dirichlet-type boundary condition, $\ub_D(\xb)$ for $\xb\in\partial\Omega$, our ultimate goal is to predict the corresponding displacement field $\ub(\xb)$, $\xb\in\omg$. 

Mathematically, let $\mcK$ be the unknown differential operator associated with the momentum balance equation which depends on the unknown tissue microstructure and mechanical properties. For a given boundary condition $\ub_D(\xb)$, the momentum balance equation and boundary conditions are
\begin{equation}\label{eqn:pde}
\begin{aligned} 
\mcK[\ub](\xb)=\mathbf{0},\quad&\xb\in \omg,\\
\ub(\xb)=\ub_D(\xb),\quad&\xb\in\partial \omg.\\
\end{aligned}
\end{equation}
Hence, our goal is to provide a surrogate solution operator for \eqref{eqn:pde} as a mapping between any arbitrary $\ub_D$ and the corresponding material response $\ub$. To this end, we propose to embrace the descriptive power of neural networks (NNs), and develop a data-driven neural operator with its input being $\ub_D(\xb)$ and its output being the displacement field $\ub(\xb)$, for any $\xb\in\omg$. Given a collection of observed function pairs $\{(\ub_D)_j(\xb),\ub_j(\xb)\}_{j=1}^N$ from DIC measurements, 
where the input $\{(\ub_D)_j\}$, $j=1,\cdots,N$ is a sequence of boundary displacement loading and  $\mcG^\dag[(\ub_D)_j](\xb)=\ub_j(\xb)$ is the corresponding (potentially noisy) displacement field. 
With neural operator learning, we aim to build an approximation of $\mcG^\dag$ by constructing a nonlinear parametric map $\mcG[\cdot\,;\,\theta]$ in the form of a NN, for some finite-dimensional parameter space $\Theta$. Here, $\theta\in\Theta$ is the set of network architecture parameters to be inferred by solving the minimization problem:
\begin{equation}\label{eqn:opt}
\min_{\theta\in\Theta}\sum_{j=1}^N\vertii{\mcG[(\ub_D)_j;\theta](\xb)-\ub_j(\xb)}_{L^2(\omg)}^2.
\end{equation}
%
In this context, we have formulated the soft tissue response modeling problem as learning the solution operator $\mcG$ of an unknown PDE system from the DIC data. 

\begin{figure*}[!ht]
	\centering
	\includegraphics[width=2.0\columnwidth]{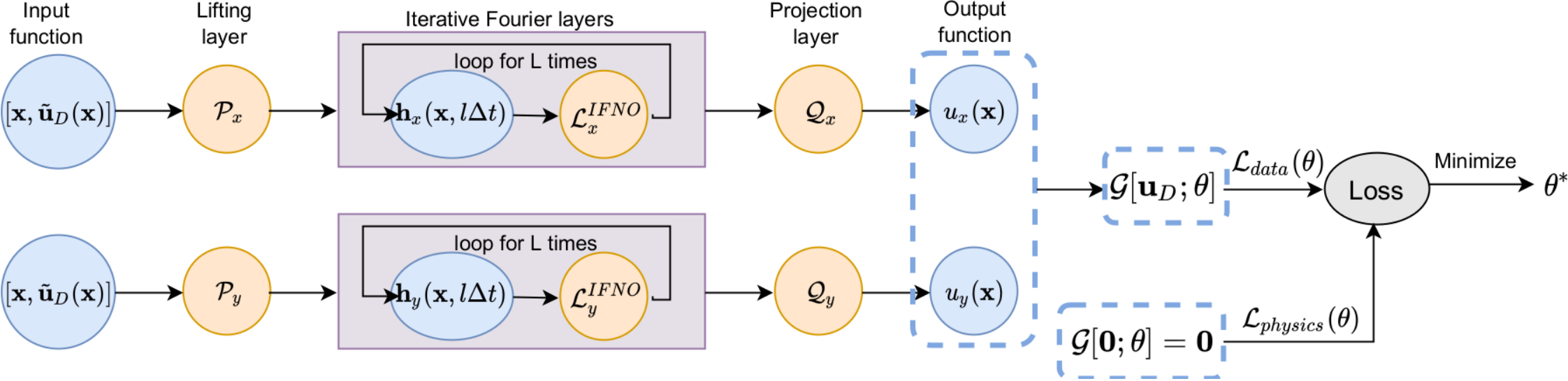}
	\caption{The architecture of the proposed physics-guided IFNO PG-IFNO), which consists of two sub-networks for the prediction of two displacement field components. Each sub-network starts from the input $[\xb,\tilde{\ub}_D(\xb)]$, then 1) Lifts to a high dimensional feature space by the lifting layer $\mathcal{P}$ and obtains the first hidden layer representation $\hb(\xb,0)$; 2) Applies $L$ iterative layers; 3) Projects the last hidden layer representation $\hb(\xb,L\Delta t)$ back to the target dimension through a shallow network $\mathcal{Q}$. The optimal network parameter $\theta^*$ is obtained by minimizing the hybrid loss function defined as the weighted sum of the data-driven loss, $\mathcal{L}_{data}$, and the physics constraint loss, $\mathcal{L}_{physics}$.}
\label{fig:domain}
\end{figure*}

Thus, our goal is to provide a {\it neural operator}, i.e., an approximated solution operator $\mcG[\cdot;\theta]:\ub_D\rightarrow \ub$, that delivers solutions of \eqref{eqn:pde} for any input $\ub_D$. Comparing with the classical PDE solvers and the NN approaches, this is a more challenging task for several reasons. First, in contrast to the classical NN approaches where the solution operator is parameterized between finite-dimensional Euclidean spaces \cite{guo2016convolutional,zhu2018bayesian,adler2017solving,bhatnagar2019prediction,khoo2021solving}, the neural operators are built as mappings between infinite-dimensional spaces 
\cite{li2020neural,li2020fourier,you2022nonlocal}. Second, for every new instance of material microstructure and/or loading scenarios $\fb$, the neural operators require only a forward pass of the network, which implies that the optimization problem \eqref{eqn:opt} \textit{only needs to be solved once and the resulting NN can be utilized to solve for new and unseen loading scenarios}. This property is in contrast to the classical numerical PDE methods \cite{leveque2007finite,zienkiewicz1977finite,karniadakis2005spectral} 
and some machine learning approaches  \cite{raissi2019physics,weinan2018deep,bar2019unsupervised,smith2020eikonet,pan2020physics}, where the optimization problem needs to be solved for every new instance of the input parameter of a known governing law. 
Finally, of fundamental importance is the fact that the neural operators can find solution maps regardless of the presence of an underlying PDE and only require the observed data pairs $\{((\ub_D)_j,\ub_j)\}_{j=1}^N$. Therefore, the neural operator learning approach is particularly promising when the mechanical responses are provided by experimental measurements, such as the displacement tracking data from DIC considered in this paper.

\subsection{Implicit Fourier neural operators (IFNOs)}\label{sec:ifno}

To provide an efficient, deep, and stable integral neural operator for the solution operator learning problem discovered above, we employ the implicit Fourier neural operators (IFNOs) \cite{you2022learning}. IFNOs stem from the idea of modeling the solution operator as a fixed point equation that naturally mimics the solution procedure for the displacement/damage fields in materials modeling. The increment between neural network hidden layers is modeled as an integral operator, which is directly parameterized in the Fourier space to facilitate the fast Fourier transformation (FFT) and accelerated learning techniques for deep networks. As shown in \cite{you2022learning}, by learning the material responses directly from data, the material microstructure and properties are learned implicitly and embedded naturally in the network parameters, enabling the prediction of the material displacement for unseen loading conditions.

    Figure \ref{fig:domain} depicts the NN architecture employed in the present work. Two IFNOs are built to predict $u_x(\xb)$ and $u_y(\xb)$, the $x$ and $y$ components of the displacement field, respectively. For each IFNO, the input is a vector function $\fb(\xb)$ on $\omg$ that contains information from $\xb$ and $\ub_D(\xb)$. Here, we notice that the displacement boundary loading $\ub_D(\xb)$ is only defined on  $\partial\omg$. To make it well-defined on the whole domain, we employ the zero-padding strategy proposed in \cite{lu2021comprehensive}, namely,  defining $\fb(\xb):=[\xb,\tilde{\ub}_D(\xb)]$ where
\begin{equation}\label{eqn:pad}
\tilde{\ub}_D(\xb)=\left\{\begin{array}{cc}
    \ub_D(\xb), & \text{ if }\xb\in\partial\omg \\
    0, &  \text{ if }\xb\in\omg\backslash\partial\omg
\end{array}\right..    
\end{equation}
Then, we lift the input $\fb(\cdot)$ to a representation (feature) $\hb(\cdot,0)$ of dimension $d$, that corresponds to the first network layer. For the consistency of notation, we label the first argument of $\hb$ as the space (the set of nodes) and the second argument as the time (the set of layers), and define the first network layer as 
$$\hb(\xb,0)=\mathcal{P}[\fb](\xb):=P\fb(\xb)+\pb,$$
where $P\in\real^{d\times 4}$ and $\pb\in\real^{d}$ are trainable parameters.

Second, we denote the $l$-th network representation by $\hb(\xb,l\Delta t)$, and formulate the NN architecture in an iterative manner: $\hb(\cdot,0)\rightarrow \hb(\cdot,\Delta t)\rightarrow\hb(\cdot,2\Delta t)\rightarrow \cdots \rightarrow \hb(\cdot,T)$, where $\hb(\cdot,j\Delta t)$, $j=0,\cdots,L:=T/\Delta t$, is a sequence of functions representing the features at each hidden layer, taking values in $\real^{d}$. Here, $l=0$ (or equivalently, $t=0$) denotes the first hidden layer, whereas $t=L\Delta t=T$ corresponds to the last hidden layer. In particular, the layer update rule in the IFNOs writes
\begin{align*}
\nonumber\hb(\xb,(l+1)&\Delta t)=\mcL^{IFNO}[\hb(\xb,l\Delta t)]\\
:=&\hb(\xb,l\Delta t)+ {\Delta t}\sigma\left(W\hb(\xb,l\Delta t)\right.\\
&\left.+\mathcal{F}^{-1}[\mathcal{F}[\kappa(\cdot;\vb)]\cdot \mathcal{F}[\hb(\cdot,l\Delta t)]](\xb)+ \cb\right).\label{eq:IFNO}
\end{align*}
Here, $\mathcal{F}$ and $\mathcal{F}^{-1}$ denote the Fourier transform and its inverse, respectively. In practice, $\mathcal{F}$ and $\mathcal{F}^{-1}$ are computed using the FFT and its inverse to each component of $\hb$ separately, with the highest modes truncated and keeping only the first $k$ modes. Also, $\cb\in\real^{d}$ defines a constant bias, $W\in\real^{d\times d}$ is the weight matrix, and $\mathcal{F}[\kappa(\cdot;\vb)]:=R\in\complex^{d\times d\times k}$ is a circulant matrix that depends on the convolution kernel $\kappa$. We further define $\sigma$ as the activation function, which is chosen to be the popular rectified linear unit (ReLU) function \cite{agarap2018deep}. Here we note that the definition of $t$ stems from the relationship established between the network update and a time stepping scheme, which enables the employment of the accelerated training strategy for the NN in the deep layer limit. 

Third, the output $u_x(\xb)$ or $u_y(\xb)$ is obtained through a projection layer. Taking the IFNO for the prediction of $u_x(\xb)$, for example, we project the last hidden layer representation $\hb(\cdot,T)$ as:
$$u_x(\xb)=\mathcal{Q}[\hb(\cdot,T)](\xb):=Q_2\sigma(Q_1\hb(\xb,T)+\qb_1)+\qb_2.$$
Here, $Q_1\in\real^{d_{Q}\times d}$, $Q_2\in\real^{1\times d_Q}$, $\qb_1\in\real^{d_Q}$, and $\qb_2\in\real$ are the trainable parameters.

Denoting the parameters and the corresponding operators associated with $u_x$ and $u_y$ with the subscripts $x$ and $y$, respectively, the vanilla version of our neural operator learning architecture without physics constraints (which will be denoted as IFNO in the following context, with a slight abuse of notation) is written as
\begin{align*}
\mcG[\ub_D;\theta](\xb):=&[\mathcal{Q}_x\circ(\mathcal{L}_x^{IFNO})^L\circ \mathcal{P}_x[\fb](\xb),\\
&~~~\mathcal{Q}_y\circ(\mathcal{L}_y^{IFNO})^L\circ \mathcal{P}_y[\fb](\xb)]\\
\approx &[u_x(\xb),u_y(\xb)]=\ub(\xb).
\end{align*}
Note that the trainable parameters are collected in  $\theta:=\{P_x,\pb_x,(Q_1)_x,(Q_2)_x,(\qb_1)_x,(\qb_2)_x,\cb_x,W_x,R_x,P_y,\pb_y,$
$(Q_1)_y,(Q_2)_y,(\qb_1)_y,(\qb_2)_y,\cb_y,W_y,R_y\}$, obtained in the vanilla IFNO by minimizing the data-driven loss only: 
\begin{equation*}
\theta^*=\underset{\theta\in\Theta}{\text{argmin}}\;\mcL_{data}(\theta),\,\,\, {\rm where}
\end{equation*}
\begin{align}
\mcL_{data}(\theta):=\sum_{j=1}^N\vertii{\mcG[(\ub_D)_j;\theta](\xb)-\ub_j(\xb)}_{L^2(\omg)}^2.\label{eqn:ifno}    
\end{align}
Further, as the layer becomes deep ($\Delta t\rightarrow 0$), the iterative architecture of the IFNOs can be seen as an analog of discretized ordinary differential equations (ODEs). This allows us to exploit the shallow-to-deep learning technique \cite{haber2018learning,modersitzki2009fair,you2022nonlocal,you2022learning}. Specifically, using the optimal network parameters $\theta^*$ obtained by training an IFNO of depth $L$, we initialize the (deeper) $\widetilde L$-layer network. As such, the optimal parameters learned on shallow networks are considered as (quasi-optimal) initial guesses for the deeper networks -- accelerating the training for deeper NNs. 

\subsection{Physics-guided neural operators}\label{sec:pi}

So far, the neural operator model introduced above fully relies on the data, and hence its predictions may not be consistent with the underlying physical principles. For instance, with the quasi-static and hyperelastic assumption of our model, the specimen has no permanent deformation. In other words, if there is no loading applied to the tissue (i.e., the specimen is at rest), we should observe zero displacement field in the specimen. However, this is generally not guaranteed in a fully data-driven neural operator model.

In this work, we aim to further leverage the neural operator learning architecture by imposing the underlying physical laws via soft penalty constraints during model training. In particular, considering a specimen at rest, the no-permanent-deformation assumption implies that zero loading should lead to zero displacement, i.e., $\mcG^\dag[\mathbf{0}]=\mathbf{0}$. To enable the neural operator predictions to be consistent with this physical constraint, we propose a {\it ``physics-guided'' neural operator model} that minimizes the residual of the above physical law together with the fitting loss from the data. This is achieved by solving the following minimization problem with a hybrid loss function: 
\begin{equation}\label{eqn:pgifno}
\theta^*=\underset{\theta\in\Theta}{\text{argmin}}\;\mcL_{data}(\theta)+\gamma \mcL_{physics}(\theta),
\end{equation}
where the data-driven loss $\mcL_{data}$ is defined in \eqref{eqn:ifno}, and the physics constraint loss $\mcL_{physics}$ is defined as
\begin{align}\label{eqn:opt_pi}
\mcL_{physics}(\theta):=\vertii{\mcG[\mathbf{0};\theta](\xb)}_{L^2(\omg)}^2.
\end{align}
Here, $\gamma>0$ is a penalty parameter to enforce the zero deformation state for a material subject to zero loading. Thus, the physics-guided neural operator is anticipated to improve the prediction performance in the small deformation regime. 
In the following, we will denote this model as the physics-guided IFNO, or the PG-IFNO, in short.

\section{Application to tissue biomechanics of the heart valve leaflet}\label{sec:setting}

We now consider the problem of learning the material response of a tricuspid valve anterior leaflet (TVAL) specimen from displacement measurements based on DIC tracking. In this problem, the constitutive equations and material microstructure are both unknown, and the dataset has unavoidable measurement noise. To demonstrate the efficacy of the proposed IFNOs in conjunction with the physics-based enrichment, we further compared our method against a conventional approach that uses constitutive modeling with parameter fittings.

\subsection{Tissue preparation and mechanical testing}\label{sec:DICexp}

In this section, we first introduce the experimental specimen and data acquisition procedure. In brief, we followed our previously established biaxial testing procedure, including acquisition of a healthy porcine heart and retrieval of the TVAL \cite{ross2019investigation,laurence2019investigation}. We then sectioned the leaflet tissue into a square specimen and measured the thickness using an optical measuring system (Keyence, Itasca, IL, USA). Afterwards, we applied a speckling pattern to the tissue surface using an airbrush and black paint \cite{zhang2004applications,lionello2014practical,palanca2016use}. The painted specimen was then mounted to a biaxial testing device (BioTester, CellScale, Waterloo, ON, Canada) with an effective testing area of $9\times9$\,mm for the following tissue characterizations (Figure  \ref{fig:dicsetup}(a)).

First, we performed a preconditioning protocol in which the specimen was subjected to 10 cycles of biaxial loading and unloading that targeted a first Piola-Kirchhoff stress of 150\,kPa to emulate the valve's \textit{in vivo} functioning conditions \cite{jett2018investigation}. Then, we performed 7 protocols of displacement-controlled testing to target various biaxial stresses: $P_{11}:P_{22}=$1:1, 1:0.66, 1:0.33, 0.66:1, 0.33:1, 0.05:1, 1:0.1, with the last two protocols for constrained uniaxial stretching in $x$ and $y$ (Figure  \ref{fig:dicsetup}(c), Table \ref{tab:dic_setting}). Here, $P_{11}$ and $P_{22}$ denote the first Piola-Kirchhoff stresses in the $x$- and $y$-directions, respectively. Each stress ratio was performed for three loading/unloading cycles. Throughout the test, images of the specimen were captured by a CCD camera, and the load cell readings and actuator displacements were recorded at 5\,Hz. Due to the use of displacement-controlled testing, we observed mild deviations from the target stresses (see Table \ref{tab:dic_setting}). 

After testing, the acquired images were analyzed using the digital image correlation (DIC) module of the BioTester's software. A $5.5\times5.5$\,mm domain in the central region of the TVAL specimen was selected since the speckling pattern was more uniform and could yield more reliable node tracking (see Figure \ref{fig:dicsetup}(a)-(b)). The pixel coordinate locations of the DIC-tracked grid were then exported for use in the subsequent study scenarios. Based on the tracked coordinates, we constructed two numerical testing datasets: (i) an original dataset obtained directly from the experimental measurements, and (ii) a smoothed dataset where moving least-squares (MLS) smoothing was performed for the nodal displacements. 

To generate the displacement fields $\ub^{original}(\xb)$ for the original samples, we subtracted each material point location with its initial location on the first sample of each protocol, and the boundary displacement was obtained by enforcing $\ub^{original}(\xb)$ on the boundary nodes.  Next, to construct the smoothed samples for the $i^{\rm th}$ material point, $\xb_i=(x_i,y_i)$, we employed a two-dimensional MLS shape function $\Psi_{i}$ to reconstruct the smoothed displacement field: $\ub^{smooth}(x_i,y_i)=\sum_{k=1}^{NP}\Psi_{k}(x_i,y_i)\ub^{original}(x_{k},y_{k})$, based on the unsmoothed displacement vector of the $NP$ points in the neighborhood of $\xb_i$. For further details regarding the MLS shape function and the smoothing procedure, we refer interested readers to  \cite{belytschko1996meshless,chen1996reproducing,you2022learning}.  Both the smoothed and the original datasets have $26,523$ total time instants (samples), denoted as $\mcD^{smooth}=\{(\ub_D)^{smooth}_j,\ub^{smooth}_j\}_{j=1}^{26,523}$ and $\mcD^{original}=\{(\ub_D)^{original}_j,\ub^{original}_j\}_{j=1}^{26,523}$, respectively. Finally, to create a structured grid for the proposed IFNOs, we further applied a cubic spline interpolation to the displacement field on a structured $21\times 21$ node grid.

\subsection{Baseline: Fung-type constitutive modeling}\label{sec:baseline}

As the baseline method for comparisons with the proposed neural operator learning methods, 
we considered a constitutive modeling approach using parameter fitting to the experimental stress-stretch data. In particular, a Fung-type model for planar stress--strain behavior was considered, with the strain energy density function given by:
$$\psi=\dfrac{c}{2}\left[\exp(a_1 E^2_{11}+a_2 E^2_{22}+2a_3 E_{11}E_{22})-1\right],$$
where $c$, $a_1$, $a_2$ and $a_3$ are the model parameters, and $E_{11}$, $E_{22}$ are the principle Green-Lagrange strains in the $x$- and $y$-directions, respectively. Based on this pre-assumed constitutive model, we aimed to find the optimal parameters ($c^*$ and $a_i^*$, $i=1,2,3$) from the training samples, and the optimal parameters were then used for predicting the displacement field on the testing samples.

In this work, constitutive model parameters were obtained by nonlinear least-squares fitting to the biaxial stress-stretch data for the training samples. In brief, the first Piola-Kirchhoff stresses in the $x$- and $y$-directions were determined using the specimen thickness $L_z$, the undeformed edge lengths $L_x$ and $L_y$, and the measured forces $F_x$ and $F_y$: $P_{11}=F_x/L_yL_z$ and $P_{22}=F_y/L_xL_z$. The two stretches were calculated as the ratio of the deformed to the undeformed edge lengths. To obtain the optimal parameters for the Fung-type model, differential evolution optimization was employed that minimizes the residual mean squared errors in the stress between the experimental data and model prediction \cite{price2006differential}. Finally, using the determined model parameters, finite element simulation was performed in Abaqus \cite{abaqus2011abaqus} with the DIC-tracked nodal displacements prescribed as boundary displacement conditions. The relative errors of displacement fields were then evaluated by comparing the finite element solution and the DIC-based measurements. In the following, we will refer to this baseline approach as the ``Fung model'' method.


\subsection{Numerical study scenarios}\label{sec:studies}

Based on the seven mechanical testing protocols listed in Table \ref{tab:dic_setting}, four study scenarios are considered to evaluate the \textit{interpolative} and \textit{extrapolative} performances of the proposed neural operator learning methods. In each study scenario, a subset of the samples were selected to form the training set and to obtain the optimal neural operator by solving \eqref{eqn:opt} and \eqref{eqn:opt_pi}. Then, the displacement field predictions were made for the remaining samples and the results were compared with the ground-truth displacement fields from the DIC measurements, to evaluate the predictivity and generalizability of our proposed methods.  Due to the relatively large number of samples, in the constitutive modeling approach it is generally intractable to perform finite element analysis for all 26,523 samples. To reduce the computational cost, we only evaluate the training and testing errors for samples in the first loading/unloading cycle of each protocol for the constitutive modeling approach. Then, we considered the averaged relative error of displacement on the first loading/unloading cycle as the error measure, so as to provide a fair comparison between the constitutive modeling and our neural operator learning approaches.

\paragraph{\textit{Study 1}} We mixed all samples from all the seven protocols, randomly selected $83\%$ of samples for training, and used the remaining for testing. In this scenario, we ensured that the boundary conditions of the samples in the testing set are inside the training region. Therefore, with this study we aimed to investigate the \textit{in-distribution} predictivity of the proposed method.

\paragraph{\textit{Study 2}} For this study, 
we employed protocols 1, 2, 4 for training and protocols 3, 5, 6 and 7 for testing.
We notice that the testing protocols are not covered in any of training sets, and they have smaller maximum tensions compared with the training sets. Hence, with this study we aimed to investigate the performance of the proposed IFNO methods for predicting the \textit{out-of-distribution material responses in the small deformation regime}. 


\begin{figure*}[h!]
  \begin{minipage}[b]{0.44\linewidth}
    \centering%
    \begin{tabular}{|>{\hspace{-4pt}}c<{\hspace{-4pt}}|>{\hspace{-4pt}}c<{\hspace{-4pt}}|>{\hspace{-4pt}}c<{\hspace{-4pt}}|>{\hspace{-4pt}}c<{\hspace{-4pt}}|}
    \hline
    \multicolumn{2}{|c|}{Model/Dataset} & Original& Smoothed\\
    \hline 
    \multirow{2}{*}{\bf Vanilla IFNO} &training error& 1.54\% & 1.43\% \\
    &testing error &{\bf 1.64\%}& {\bf 1.53\%}\\
    \hline 
    \multirow{2}{*}{Fung model} &training error &  10.34\%&10.10\%\\
    &testing error &  10.83\%&10.53\% \\
    \hline
    \end{tabular}
    \par\vspace{60pt}
      \end{minipage}
      \begin{minipage}[b]{0.55\linewidth}
    \centering
\subfigure{\includegraphics[width=1.\columnwidth]{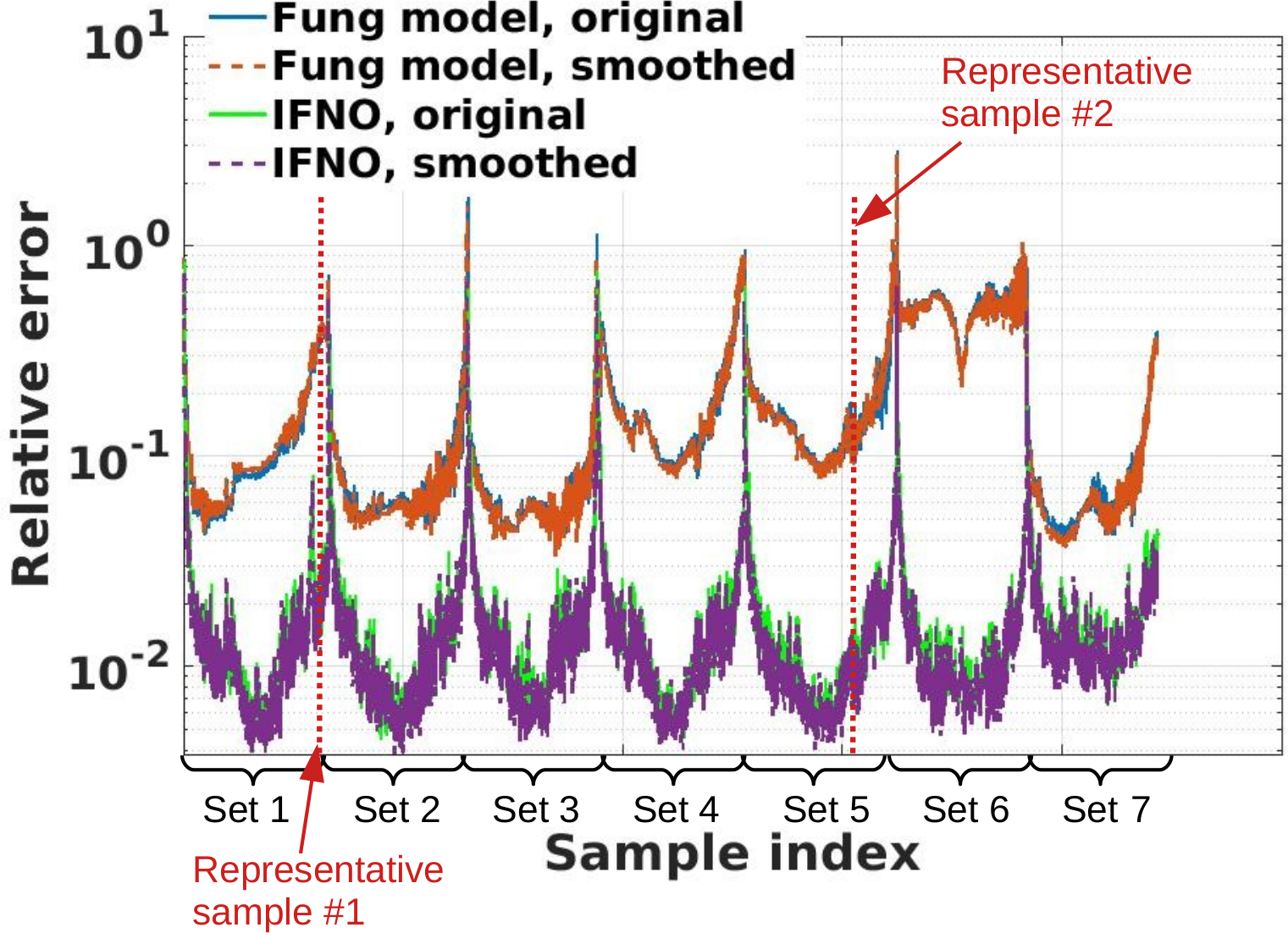}}
    \par\vspace{0pt}
  \end{minipage}%
\caption{Error comparisons of different models -- Study 1: {\it in-distribution} prediction with training set on 83\% of randomly selected samples. {\it Left}: relative errors for the displacement field prediction on the training and test datasets. We highlight the model with the best prediction accuracy in bold. {\it Right}: sample-wise error comparison on all biaxial testing protocol sets.\vspace{15pt}}
    \label{fig:loss_study1}
\end{figure*}

\begin{figure*}[h!]
    \centering
    \subfigure{\includegraphics[width=1.\textwidth]{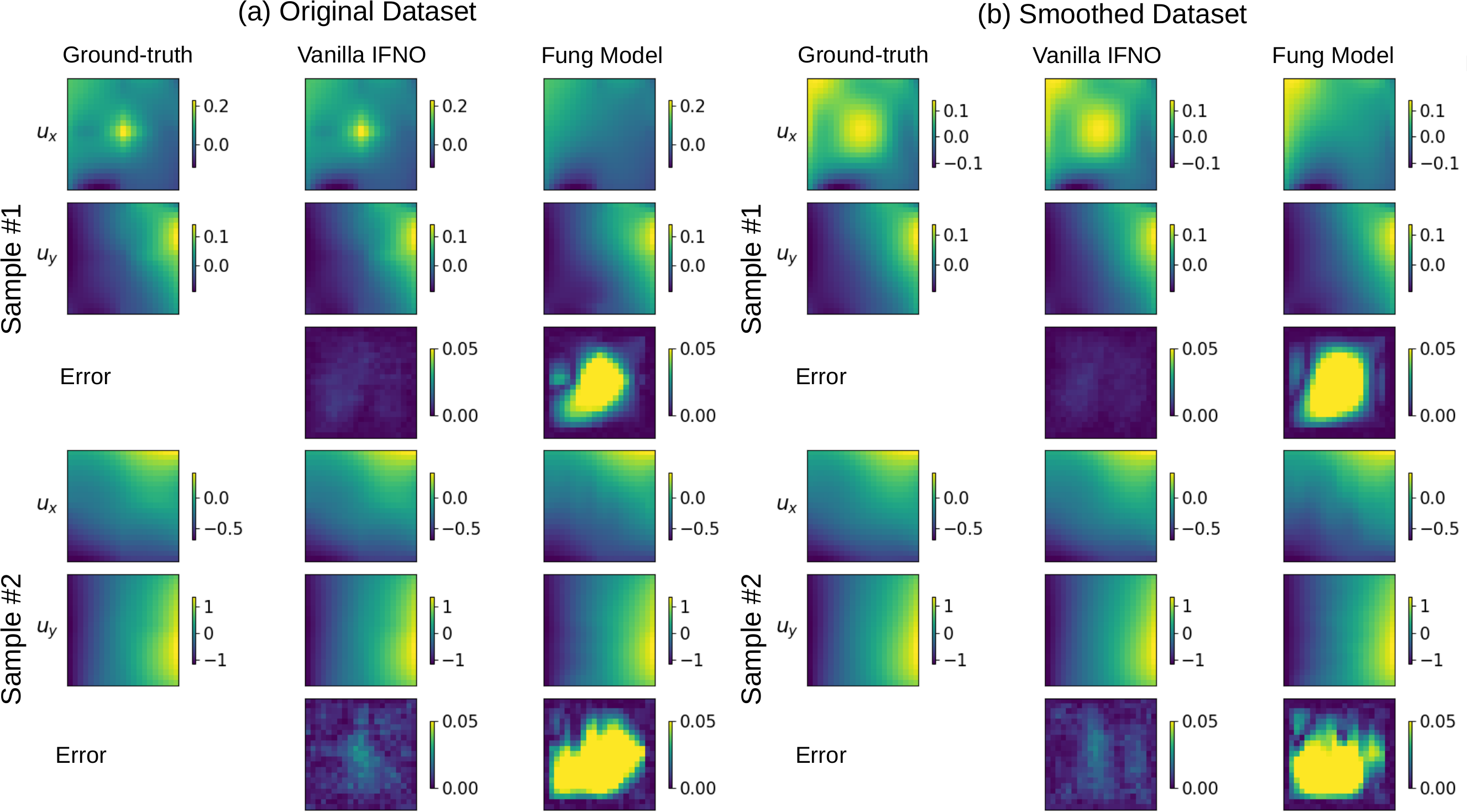}}
    \caption{Visualization of the Fung-type model fitting and IFNO performances on two representative test samples in (a) the original dataset and (b) the smoothed dataset -- Study 1. (Correspond the two representative test samples highlighted in Figure \ref{fig:loss_study1} -- {\it right}.)}
    \label{fig:unsmoothdata_study1}
\end{figure*}

\paragraph{\textit{Study 3}} 
We used protocols 1, 6, and 7 for training and protocols 2-5 for testing. 
The protocols considered in testing were not covered in any of the training protocols, although the deformation range of the testing protocols may fall inside the range of the training ones. Hence, we attempted to illustrate the {\it out-of-distribution prediction on the intermediate deformation regime}.

\paragraph{\textit{Study 4}} We used protocols 2-7 for training, and protocol 1 for prediction. We notice that the equibiaxial tension protocol ($P_{11}:P_{22}=$1:1) is not covered in any of other sets, and protocol 1 exhibits the largest maximum tensions among all the sets. Hence, with this study, we aimed to investigate the \textit{out-of-distribution predictivity in the large deformation regime} of the proposed method. 

\section{Results and Discussions}\label{sec:results}

In this section, we illustrate the performance of the proposed neural operator learning approaches. All our numerical experiments were performed on a machine with 2.8 GHz 8-core CPU and a single Nvidia RTX 3060 GPU, using a Pytorch implementation modified from the package provided in \cite{li2020fourier}. The optimization was performed with the Adam optimizer. For all IFNOs, we set the dimension of $\hb$ as $d=16$ and the number of truncated Fourier modes as $k=8\times 8$, with $L=12$ hidden layers. The network was trained with the shallow-to-deep training procedure: we initialized the $L-$layer network parameters from the $(L/2)-$layer IFNOs model. For each depth $L$, we trained the network for 1,000 epochs with a learning rate of $3\times10^{-3}$, then decreased the learning rate with a ratio of $0.5$ every 100 epochs. For all PG-IFNOs we took the penalty parameter $\gamma=1.0$, although we noted that this parameter can be further hand-tuned or optimized to potentially achieve a better performance.

\begin{figure*}[h!]
  \begin{minipage}[b]{0.45\linewidth}
    \centering%
    \begin{tabular}{|>{\hspace{-4pt}}c<{\hspace{-4pt}}|>{\hspace{-4pt}}c<{\hspace{-4pt}}|>{\hspace{-4pt}}c<{\hspace{-4pt}}|>{\hspace{-4pt}}c<{\hspace{-4pt}}|}
    \hline
    \multicolumn{4}{|>{\hspace{-4pt}}c<{\hspace{-4pt}}|}{Train on sets 1, 2, 4, test on sets 3, 5, 6, 7}\\
    \hline
    \multicolumn{2}{|c|}{Model/Dataset} & Original & Smoothed \\
    \hline
    \multirow{3}{*}{Vanilla IFNO} & training error & 1.53\% & 1.51\% \\
    & testing error &  16.78\%&18.80\% \\
    \hline 
    \multirow{3}{*}{\bf PG-IFNO} & training error& 1.51\% & 1.50\%\\
    & testing error &  {\bf 15.32\%}&15.76\% \\
    \hline 
    \multirow{3}{*}{Fung model} & training error &  12.37\%&12.54\%\\
    & testing error &  16.80\%&{\bf 15.70\%} \\
    \hline
    \end{tabular}
    \par\vspace{30pt}
      \end{minipage}
      \begin{minipage}[b]{0.55\linewidth}
    \centering
\subfigure{\includegraphics[width=1.0\columnwidth]{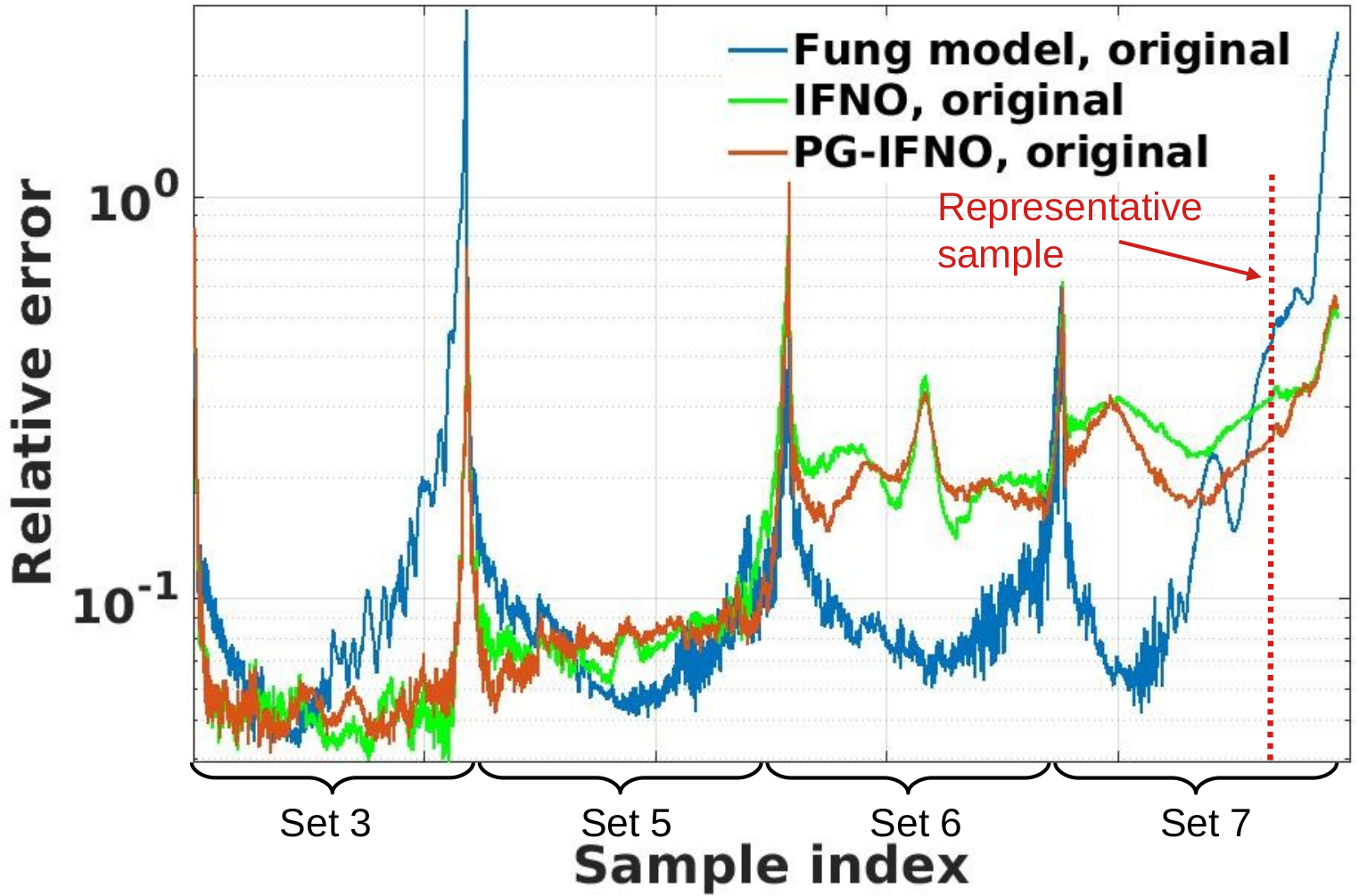}}
    \par\vspace{0pt}
  \end{minipage}%
\caption{Error comparisons of different models -- Study 2: {\it out-of-distribution} prediction on the {\it small} deformation regime. {\it Left}: relative errors for the displacement field prediction on the training and test datasets. We highlight the model with the best prediction accuracy in bold. {\it Right}: sample-wise error comparison on all test sets from the original (unsmoothed) dataset.}
    \label{fig:loss_study2}
\end{figure*}

\begin{figure*}[h!]
    \centering
    \subfigure{\includegraphics[width=1.0\textwidth]{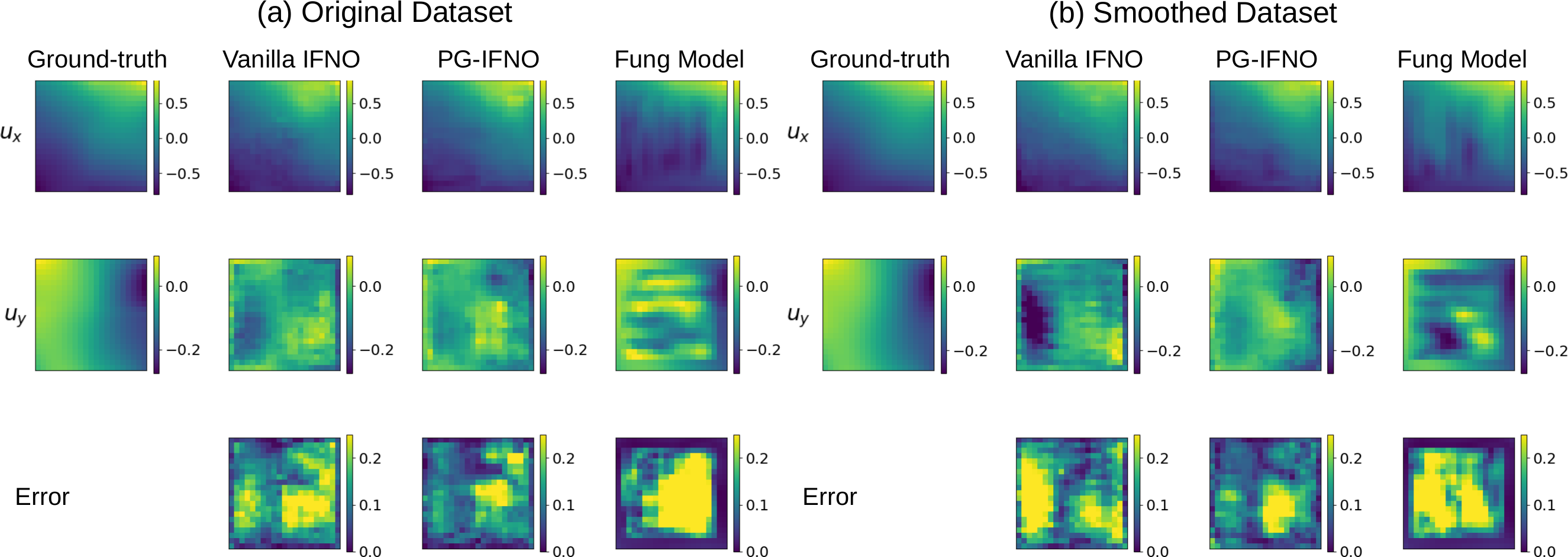}}
    \caption{Visualization of the Fung-type model fitting, IFNO and PG-IFNO performances on a test sample in (a) the original dataset and (b) the smoothed dataset -- Study 2. (Corresponds to the representative test sample defined in Figure \ref{fig:loss_study2} -- {\it right}).}
    \label{fig:unsmoothdata_study2}
\end{figure*}

\paragraph{Study 1: In-distribution prediction} To verify the model's predictivity for in-distribution learning tasks, in this study we randomly selected $83\%$ of the samples of all protocols to form the training set, and then built the vanilla IFNO model and the Fung-type model based on this common training set. Figure \ref{fig:loss_study1} ({\it left}) shows the relative displacement errors from both the original and smoothed datasets, and the sample-wise errors for each model are provided in Figure \ref{fig:loss_study1} ({\it right}). When comparing the results between the original dataset and the smoothed dataset, one can observe that the smoothing procedure slightly improves the prediction accuracy for both the INFO and Fung-type models. By comparing the prediction accuracy between the two computational models, the IFNO outperforms the conventional constitutive modeling approaches by around one order of magnitude, on both the original and smoothed datasets. To provide further insights into this comparison, in Figure  \ref{fig:unsmoothdata_study1} we visualized both the {\it x}- and {\it y}-displacement solutions and the prediction errors obtained with the IFNO and the Fung-type model on two test samples, which correspond to the large deformation (sample \#2) and small deformation (sample \#1) representatives, respectively. The Fung-type model, which considered the homogenized stress--strain at one material point (i.e., the center of the specimen) due to limited information about the spatial variation in the stress measurement,  failed to capture the material heterogeneity and hence exhibited large prediction errors in the interior region of the TVAL specimen domain. This observation confirms the importance of capturing the material heterogeneity and verifies the capability of the IFNOs in heterogeneous material modeling.

\paragraph{Study 2: Out-of-distribution prediction on the small deformation regime} In this study, three protocols with the largest tensions (i.e., sets 1, 2, and 4) were used for training, and the other four protocols were used for prediction validation (sets 3, 5, 6, and 7 as listed in Table \ref{tab:dic_setting}). Since the prediction sets are with a different biaxial loading ratio that is unseen from the training samples, this is an {\it extrapolative} learning task in the small deformation region. Figure \ref{fig:loss_study2} ({\it left}) provides the relative displacement errors from all models. One can see that comparing with the interpolative prediction task in Study 1, the extrapolative predictions are less effective for both the neural operator and the constitutive models. It was in particular noted that for the vanilla IFNO model while the training error was at a relatively low error (i.e., the model still possessed good expressivity in sets 1, 2, and 4), the testing error deteriorates by 10 times and reached a similar level to the constitutive modeling error. Perhaps unsurprisingly, this observation again verifies the sensitivity of machine learning models in extrapolative tasks (see, e.g., \cite{he2021manifold}). As shown in Figure \ref{fig:loss_study2} ({\it right}), we demonstrate the sample-wise errors from the original dataset for each model, and we noticed that the results on the smoothed dataset exhibit a similar trend. One can observe that for the Fung-type model the level of prediction errors is relatively similar for all four testing sets, while large errors are observed in sets 6 and 7 (the sets with the smallest maximum tensions) for the vanilla IFNO model. Those sets are the furthest away from the training set and hence their sample distributions are substantially different from those in the training sets. 

\begin{figure*}[h!]
  \begin{minipage}[b]{0.45\linewidth}
    \centering%
    \begin{tabular}{|>{\hspace{-4pt}}c<{\hspace{-4pt}}|>{\hspace{-4pt}}c<{\hspace{-4pt}}|>{\hspace{-4pt}}c<{\hspace{-4pt}}|>{\hspace{-4pt}}c<{\hspace{-4pt}}|}
    \hline
    \multicolumn{4}{|>{\hspace{-4pt}}c<{\hspace{-4pt}}|}{Train on sets 1, 6, 7, test on 2-5}\\
    \hline
    \multicolumn{2}{|c|}{Model/Dataset} & Original & Smoothed \\
    \hline
    \multirow{2}{*}{\bf Vanilla IFNO} & training error & 1.51\% & 1.43\% \\
    & testing error &{\bf 8.03\%}& {\bf 8.09\%}\\
    \hline 
    \multirow{2}{*}{PG-IFNO} & training error &  1.47\%& 1.35\%\\
    & testing error & 8.75\% & 9.08\%\\
    \hline 
    \multirow{2}{*}{Fung model} & training error &  15.49\%&14.48\%\\
    & testing error &11.75\%& 11.33\%\\
    \hline
    \end{tabular}
    \par\vspace{40pt}
      \end{minipage}
      \begin{minipage}[b]{0.55\linewidth}
    \centering
    \par\vspace{0pt}
  \end{minipage}%
      \begin{minipage}[b]{0.5\linewidth}
    \centering
\subfigure{\includegraphics[width=1.\columnwidth]{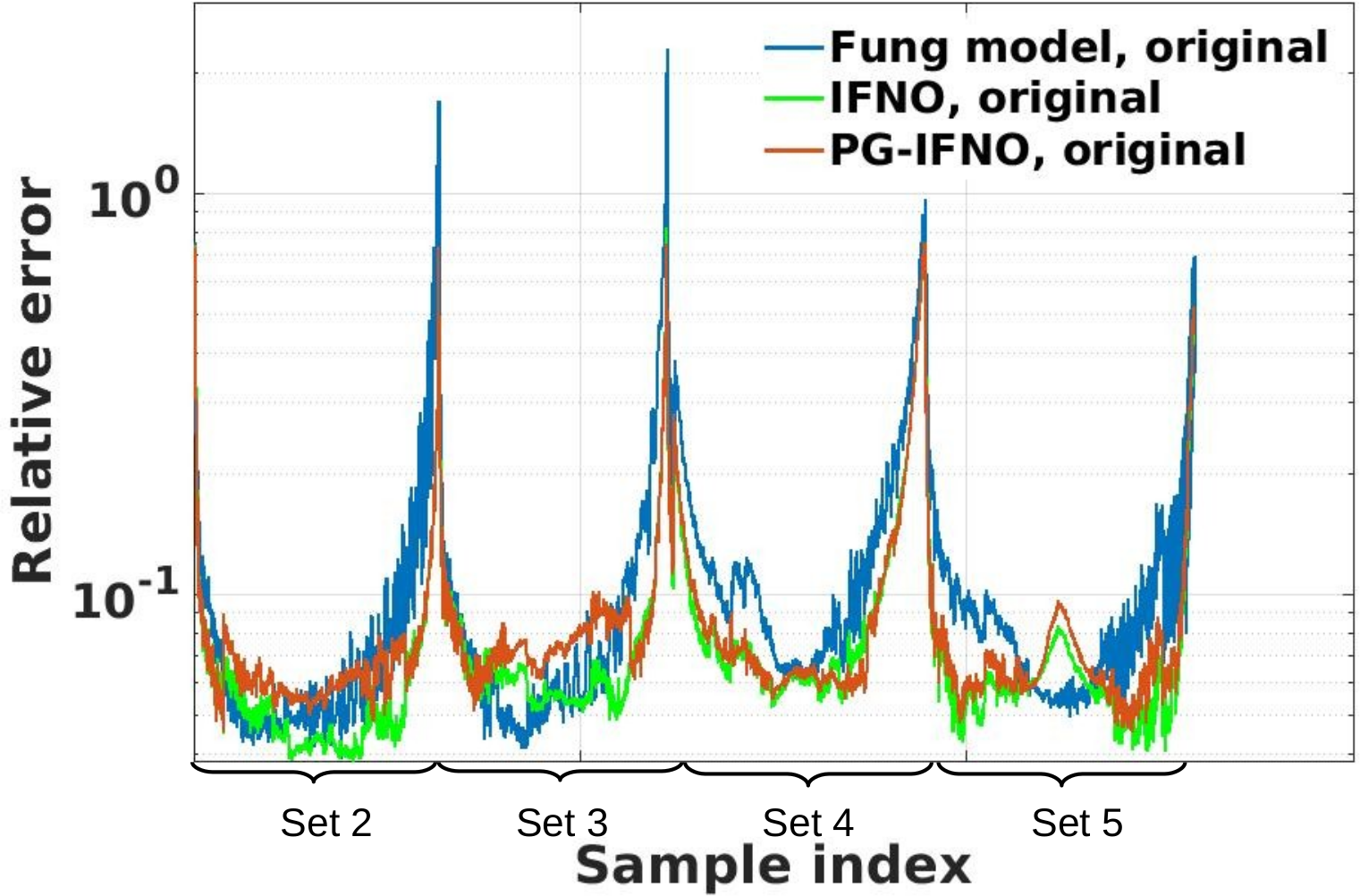}}
    \par\vspace{0pt}
  \end{minipage}%
\caption{Error comparisons of different models -- Study 3: {\it out-of-distribution} prediction on the intermediate deformation regime. {\it Left}: relative errors for the displacement field prediction on the training and test datasets. We highlight the model with the best prediction accuracy in bold. {\it Right}: sample-wise error comparison on all test sets from the original (unsmoothed) dataset.}
    \label{fig:loss_study3}
\end{figure*}

In this study, we also investigated the performance of the proposed PG-IFNOs. By infusing the no-permanent-deformation constraint, an improvement of the testing error was observed. On the original dataset, the PG-IFNO has outperformed the Fung-type model by 1.5\%, and achieves an on-par performance on the smoothed dataset as well. From the sample-wise errors, we can see that the PG-IFNOs generally improved the prediction accuracy on sets 6 and 7 -- which are the protocols on the small deformation regime. This fact is also verified by the solutions and prediction errors on a representative test sample in set 7, as depicted in Figure \ref{fig:unsmoothdata_study2}. Comparing with the original IFNO, the PG-IFNO has obtained visually more consistent predictions on the component with constraint tension. Hence, these results suggest that sufficient coverage of sample distribution in the training protocol is critical for neural operator learning methods, while it has less impact for the constitutive modeling approach if the pre-defined model form exhibits good generalizability like the adopted Fung-type model. On these challenging extrapolative learning tasks, incorporating proper physics constraints seems to make the neural operator learning more versatile.

\paragraph{Study 3: Out-of-distribution prediction on the intermediate deformation regime} In this study, protocol sets 1, 6, and 7 were used in model training, while the rest of sets (protocols 2-5) were for prediction validation. As such, the prediction sets are still with unseen tension ratios from the training sets, but the deformation range of the testing protocols are {\it within} the range of the training ones. In Figure \ref{fig:loss_study3} ({\it left}),  the relative displacement errors are provided. We can see that the testing errors from the IFNOs are still much larger than their respective training errors, due to the out-of-distribution learning nature of this study. However, the prediction error from the IFNOs outperforms the Fung-type model by around 3.5\%. Therefore, as long as the dataset has sufficient coverage of the deformation, neural operator learning methods can be a more effective approach than the traditional constitutive models, even in the more challenging out-of-distribution tasks. To provide further insights into this comparison, Figure \ref{fig:loss_study3} ({\it right}) depicts the sample-wise errors on the original (unsmoothed) dataset. One can see that the prediction errors are almost uniform among all protocol sets. In this study, the PG-IFNO does not substantially improve the prediction accuracy compared with the original IFNO, possibly because the physics constraint stems from the zero loading case and is more helpful on guiding the prediction in {\it small} deformation regimes.

\begin{figure*}[h!]
  \begin{minipage}[b]{0.45\linewidth}
    \centering%
    \begin{tabular}{|>{\hspace{-4pt}}c<{\hspace{-4pt}}|>{\hspace{-4pt}}c<{\hspace{-4pt}}|>{\hspace{-4pt}}c<{\hspace{-4pt}}|>{\hspace{-4pt}}c<{\hspace{-4pt}}|}
    \hline
    \multicolumn{4}{|>{\hspace{-4pt}}c<{\hspace{-4pt}}|}{Train on sets 2-7, test on 1}\\
    \hline
    \multicolumn{2}{|c|}{Model/Dataset} &Original&Smoothed\\  
    \hline
    \multirow{2}{*}{Vanilla IFNO} &training error& 1.43\% &  1.38\%\\
    &testing error& 13.07\% & 14.05\% \\
    \hline 
    \multirow{2}{*}{PG-IFNO} & training error & 2.08\% & 1.36\%\\
    & testing error & 15.90\% & 14.54\%\\
    \hline 
    \multirow{2}{*}{\bf Fung model} &training error&  15.07\%&14.26\%\\
    &testing error&  {\bf 10.34\%}&{\bf 10.40\%} \\
    \hline
    \end{tabular}
    \par\vspace{40pt}
      \end{minipage}
      \begin{minipage}[b]{0.5\linewidth}
    \centering
\subfigure{\includegraphics[width=1.\columnwidth]{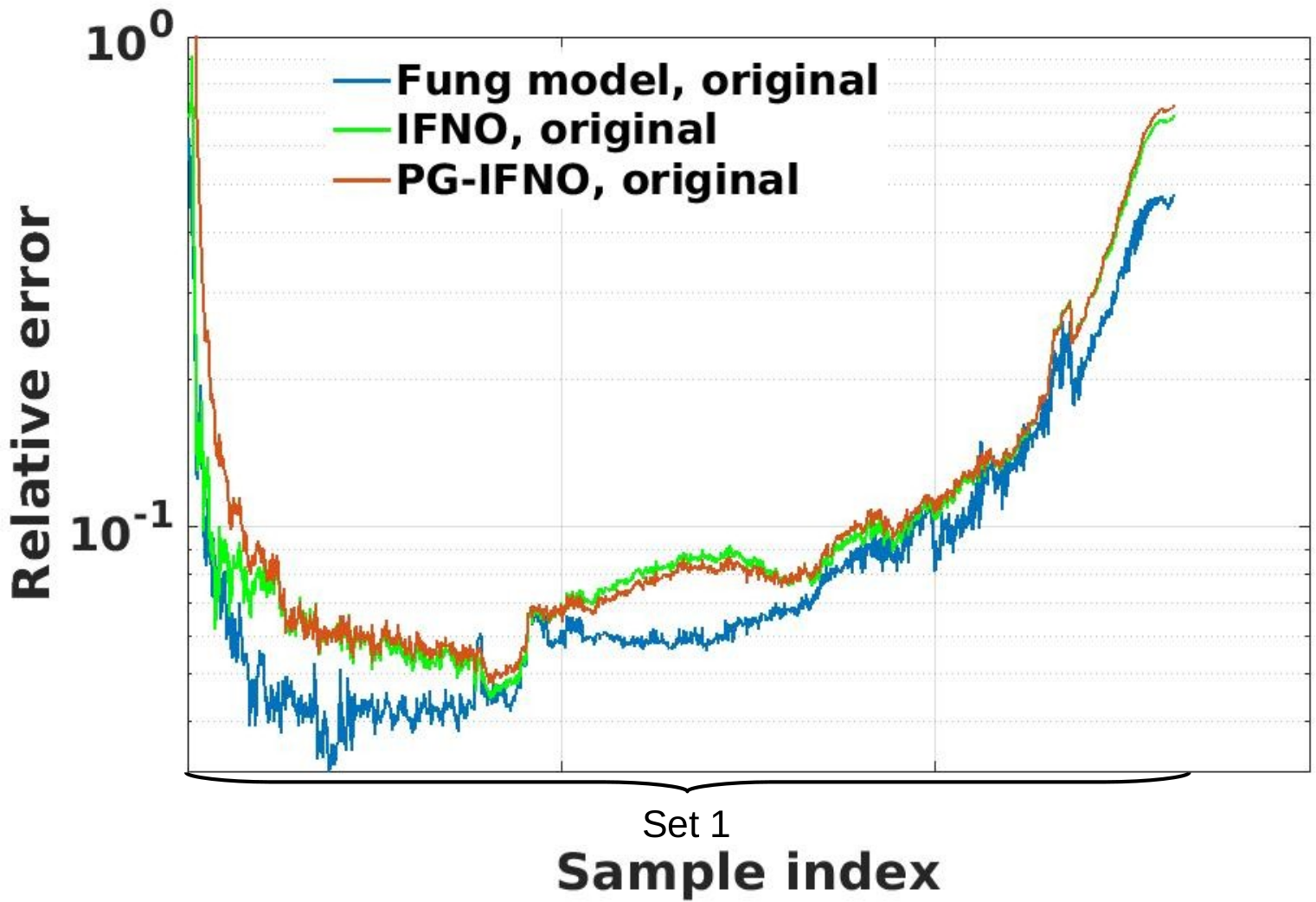}}
    \par\vspace{0pt}
  \end{minipage}%
\caption{Error comparisons of different models -- Study 4: {\it out-of-distribution} prediction on the {\it large} deformation regime. {\it Left}: relative errors for displacement field prediction on the training and test datasets. We highlight the model with the best prediction accuracy in bold. {\it Right}: sample-wise error comparison on all test sets from the original (unsmoothed) dataset.}
    \label{fig:loss_study4}
\end{figure*}

\paragraph{Study 4: Out of distribution prediction on the large deformation regime} In the last comparative study scenario, the experimental protocol for prediction was selected to be the one with the largest maximum tension, with the aim to evaluate the models' extrapolative prediction abilities on the large deformation regime. From Figure \ref{fig:loss_study4}, one can see that while the IFNO becomes less effective on both datasets, the Fung-type model exhibits a better fit to both the original and smoothed testing datasets. On the other hand, the physics constraint does not help much on this study. These results suggest that to ensure reliable predictions from neural operator learning methods, with an insufficient coverage of deformation range in the training protocols, a judiciously designed physics constraint for the data range becomes important.

\section{Conclusion}\label{sec:conclusion}

In this work, we have applied the neural operator learning method to modeling the mechanical responses of a biological tissue specimen under different loading conditions. Specifically, a data-driven computing workflow has been proposed, which learns the material model directly from the DIC displacement tracking measurements and integrates material identification, modeling procedures, and material responses prediction into one unified learning framework. With the proposed neural operator learning, the mechanical response of this tissue specimen can be modeled as a data-driven solution operator from the boundary loading to the resultant displacement field, and the learnt model will be applicable to unseen loading conditions. To verify its efficacy on real-world soft tissue response learning tasks which feature spatial heterogeneity, measurement noise, anisotropic and nonlinear behaviors, we have used the proposed workflow to model a porcine heart TVAL specimen with the DIC measurement data collected from biaxial and constrained uniaxial tension tests. In the in-distribution validation case, our proposed neural operator learning method has been shown to significantly outperform the conventional constitutive modeling approach, with its predictions on out-of-distribution learning tasks being less effective. To improve the model generalizability on out-of-distribution tasks, we have further leveraged the neural operator learning method towards physically-consistent predictions for tissue at rest, and proposed a physics-guided neural operator learning approach. Numerical studies have demonstrated substantial improvements in terms of enhanced generalization performance in the small deformation regime. Hence, these results suggest that with a sufficient coverage of training sample distribution and/or properly designed physics constraints, the neural operator learning approach could offer an alternative approach which outperforms the conventional phenomenological model in complex and heterogeneous material modeling tasks.

Despite the encouraging results presented herein, numerous questions and potentials require further investigations. For example, although the proposed no-permanent-deformation constraint seems effective in the small deformation regime, it has little impacts on improving the prediction accuracy on the large deformation regimes. As a natural extension, we will consider the enforcement of other physics constraints, which would potentially further enhance the performance of the method's {\it extrapolative} predictivity. On the other hand, another important next step is to consider other boundary loading scenarios in our learning framework, such as the traction loading problems. Finally, another question arises from the possibility of achieving the improved performance by optimizing the penalty parameter $\gamma$ in the physics-guided hybrid loss function \eqref{eqn:pgifno}. It has been shown that an optimized penalty parameter could further enhance the accuracy and trainability of the constrained neural networks (see, e.g., \cite{wang2021learning,wang2022and,mcclenny2020self}). Hence, the performance of PG-IFNO might get further enhanced by designing effective algorithms which selects appropriate weights in the hybrid loss function.

\begin{acknowledgment}
H. You and Y. Yu would like to acknowledge support from the National Science Foundation under award DMS 1753031. Portions of this research were conducted on Lehigh University's Research Computing infrastructure partially supported by NSF Award 2019035. C. Ross, C.-H. Lee and M.-C. Hsu would like to acknowledge support from the Presbyterian Health Foundation Team Science Grant. C. Ross was in part supported by the National Science Foundation Graduate Research Fellowship Program (GRF2020307284). M.-C. Hsu was in part supported by the National Institutes of Health under award number R01HL142504.
\end{acknowledgment}




\bibliographystyle{asmems4}

\bibliography{snl}





\end{document}